\documentclass[conference]{IEEEtran}
\IEEEoverridecommandlockouts
\usepackage{cite}
\usepackage{amsmath,amssymb,amsfonts}
\usepackage{algorithmic}
\usepackage{graphicx}
\usepackage{textcomp}
\usepackage{xcolor}
\usepackage[normalem]{ulem}
\usepackage{float}
\usepackage{caption}
\usepackage{graphicx}
\usepackage{tikz}
\usetikzlibrary{shadows,decorations.pathreplacing,positioning,calc,arrows.meta}
\usepackage{pgfplots}
\pgfplotsset{compat=1.17}
\newcommand\copyrighttext{%
    \scriptsize \textcopyright 2026 IEEE. Personal use of this material is permitted.
  Permission from IEEE must be obtained for all other uses, in any current or future
  media, including reprinting/republishing this material for advertising or promotional
  purposes, creating new collective works, for resale or redistribution to servers or
  lists, or reuse of any copyrighted component of this work in other works.}
\newcommand\copyrightnotice{%
    \pubid{\parbox{\textwidth}{\centering \copyrighttext}}
}
\makeatletter
\def\@pubidpullup{2.5\baselineskip}
\makeatother

\title{CommandLM: Data driven behavior level descriptor for ego vehicles}

 \author{
 	\parbox{\textwidth}{%
 		\centering
 		Boris Tokic$^{1}$, Constantin Selzer$^{1}$, Fabian B. Flohr$^{2}$%
 	}%
 	\thanks{$^{1}$Munich University of Applied Sciences, Munich, Germany
 		{\tt\small tokic@hm.edu, constantin.selzer@hm.edu}}%
 	\thanks{$^{2}$Munich University of Applied Sciences, Munich, Germany
 		{\tt\small fabian.flohr@hm.edu}}%
 }
\begin{document}
\copyrightnotice
\maketitle

\begin{abstract}
    As autonomous driving systems move toward real-world deployment, interpretable, behavior-level decision-making is essential for safety, trust, and regulation. We introduce CommandLM, a multimodal large language model that generates concise, human-readable behavior descriptions for ego vehicles from fused multi-sensor data. Our model processes temporally fused bird’s-eye view representations from LiDAR and multi-camera inputs via a Q-Former adapter connected to a quantized, LoRA-fine-tuned large language model. Trained on our CommandLM-nuScenes dataset, CommandLM produces intent-aware, interpretable captions suitable for planner supervision and safety auditing. Experiments demonstrate strong linguistic and behavioral alignment, achieving CIDEr 0.67, and BERT-F1 0.88, substantially outperforming the BLIP-2 baseline (CIDEr 0.52, BERT-F1 0.86). In human evaluation, 58\% of the generated descriptions were rated accurate, efficient and rule-compliant, confirming their real-world plausibility. While the remaining descriptions may not always select the most efficient, goal-oriented behavior, CommandLM’s interpretable outputs enable downstream validation systems to identify and correct such cases, making it an effective tool for transparent behavior auditing. These results show that integrating multimodal fusion with language reasoning yields efficient and transparent behavior-level understanding for autonomous driving. We release our code and dataset at: https://github.com/b-tok/CommandLM
\end{abstract}


\section{Introduction}
\noindent
Advancing autonomous driving systems to achieve human-level scene understanding and explainability has become increasingly critical as these technologies move toward widespread deployment~\cite{DrivingWithLLM}. A central challenge is not simply perceiving the environment, but intelligently interpreting complex, dynamic traffic scenes and conveying intentions in a transparent, human-comprehensible manner. Interpretable behavior-level reasoning enables autonomous vehicles to justify their actions to developers, passengers, and regulators, supports improved safety audits, and fosters public trust~\cite{Atakishiyev2021}. Such capabilities also enable next-generation driving assistants and robust downstream planners through natural, explainable supervision.
Despite steady progress in fusing multi-modal sensor data and applying large language models (LLMs) to robotics, translating raw perception into high-level, context-aware driving behaviors remains a challenge~\cite{Chen2024}. Autonomous vehicles must synthesize high-dimensional LiDAR and camera streams, comprehend interactions among diverse agents, infer intent, and plan around ambiguous cues, all while compressing this complexity into actionable and interpretable summaries. Naive approaches that map directly from sensors to control signals often sacrifice transparency and interpretability, while rule-based explanations fail to generalize to real-world variability and dynamic traffic conditions~\cite{Chen2024, Wen2025}. End-to-end LLM-driven controllers face significant computational demands that hinder real-time deployment on resource-constrained vehicle platforms, and models trained on coordinate-based representations struggle to capture abstract scene dynamics and behavior-level intent due to the inherent numerical insensitivity of LLMs~\cite{Wen2025, Acharya2025, Cui2025Survey, chen-etal-2023-improving}.

\begin{figure}[H]
  \centering
  \includegraphics[width=0.48\textwidth]{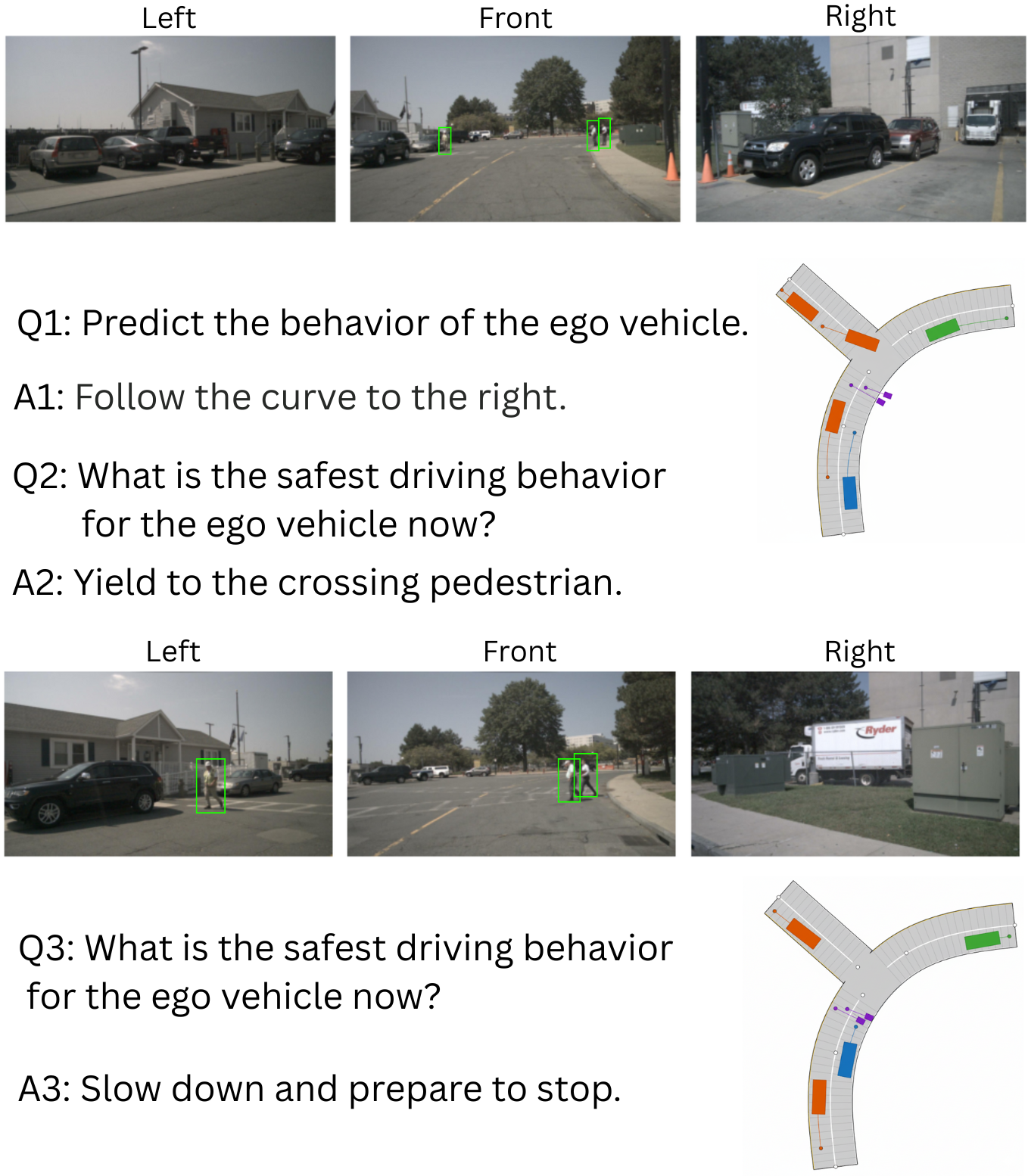}
  \caption{Sequential decision-making by CommandLM. The ego vehicle in blue adapts from continuing straight in the first sample to yielding in the second sample as pedestrians in purple cross the road.}
  \label{fig:pedestrian}
\end{figure}

\pubidadjcol

\noindent
CommandLM addresses these challenges with a multimodal large language model
(MLLM) architecture tailored for behavior-level ego-vehicle description.
The proposed system, CommandLM, processes temporally fused BEV representations from multi-camera and LiDAR data, distilling each driving scenario into concise, intent-driven textual captions interpretable by humans. We perform studies on how to best align data, architecture, and training strategy to yield generalizable behavior-level outputs that aid both human understanding and downstream planning modules, as illustrated by a short example in Fig.~\ref{fig:pedestrian}.
Previous studies have explored methods such as incorporating absolute coordinates into linguistic components and passing large volumes of visual data to LLMs, which can lead to excessive computational demands~\cite{DriveLM, DriveVLM}. These strategies are incompatible with the approach proposed here.
While OmniDrive~\cite{OmniDrive} uses a temporally aware architecture for end-to-end perception and planning, this work applies temporal context to a different goal, behavior-level scene understanding based on question-answering (QA) driven supervision, emphasizing interpretability over control.

The contributions of our approach are twofold:
\begin{enumerate}
  \item Introduction of a novel architecture that fuses dense BEV representations from multi-camera and LiDAR inputs with temporal buffering and couples them with a Q-Former adapter for a quantized, LoRA-tuned LLM, enabling salient query extraction, rich scene encoding, and interpretable, intent-driven temporal reasoning.
  \item Development of a dataset augmentation pipeline that generates object-centric, behavior-focused, and linguistically diverse supervision, reducing reliance on coordinate-heavy or brittle cues and improving generalization.
\end{enumerate}

\noindent
The proposed approach assumes access to synchronized multi-sensor data with sufficient scene diversity and acknowledges that behavior-level inference is constrained to the breadth of scenarios present in the training corpus.
In summary, we present and systematically investigate a practical path to scalable, explainable autonomous driving by harnessing recent advances in MLLMs, optimized training, and dataset engineering to bridge the gap between raw perception and transparent decision-making.

\section{Related Work}
\noindent
The Transformer’s success in NLP catalyzed its adaptation to vision tasks, most notably in Vision Transformers, which treat images as token sequences~\cite{ViTs}. This foundation enabled powerful multimodal models by integrating vision and language encoders. OpenAI’s CLIP~\cite{CLIP} demonstrated that jointly training image and text encoders on vast internet data yields strong, aligned representations for zero-shot tasks, though CLIP lacks generative abilities and specialized outputs (like image captions).
BLIP-2~\cite{BLIP2} resolves these shortcomings through its Q-Former module, which aligns frozen visual features with LLMs via 32 learnable queries. Q-Former’s efficiency and adaptability make it popular for autonomous driving, with several applications in this context~\cite{DriveLM,OmniDrive, LMDrive}.
%

\textbf{Recent works on Visual Question Answering (VQA) in Autonomus Driving} explore LLMs as core modules in self-driving systems. LMDrive~\cite{LMDrive} pioneers end-to-end control by aligning multi-view camera and LiDAR features with an LLM, using Q-Former for feature extraction, with evaluation conducted on the CARLA simulator. DriveLM~\cite{DriveLM} takes a complementary approach, leveraging BLIP-2 to perform VQA on simulated and real-world data. The work demonstrates that language-based reasoning can effectively model complex driving behavior from sensor data.
EMMA~\cite{EMMA} frames all driving tasks as VQA, mapping directly from camera images to outputs like trajectory or perception objects in natural language. It achieves state-of-the-art results on nuScenes and competitive performance on Waymo, but processes only a few frames at a time, lacks LiDAR integration, and is not open-source.
OmniDrive~\cite{OmniDrive} advances the field by modeling 3D situational awareness, compressing video into a 3D world model for the LLM to support robust planning and reasoning.
These works establish that MLLMs can serve as both direct controllers and reasoning engines enhancing perception and planning interpretability.
%
%
%

\textbf{Several studies on real-time capabilities of LLMs in Autonomous Driving} have explored balancing LLM reasoning with latency constraints. AsyncDriver~\cite{AsyncDriver} addresses this through asynchronous inference, updating features for a high-frequency planner every few frames. LeapAD~\cite{LeapAD} employs a dual-process approach, using fast heuristics for routine driving and slower reasoning for complex scenarios with a memory bank for adaptation.
Recent work formulates motion prediction as sequence modeling: MotionLM~\cite{MotionLM} discretizes trajectories into tokens for autoregressive prediction without anchors or latent variables, while VisionTrap~\cite{VisionTrap} incorporates visual cues from multi-camera input with language model supervision during training, achieving 53 ms latency during inference. Demonstrating that language-guided training can enhance both prediction accuracy and interpretability without sacrificing real-time responsiveness.
%

\textbf{High-quality VQA datasets are essential for interpretable autonomous driving.} BDD-X~\cite{BDD-X} pioneered this direction with action-explanation pairs, providing front camera images and textual driver justifications. 
DriveLM~\cite{DriveLM} advances this by introducing Graph VQA, providing multi-view camera images, LiDAR point clouds, and interconnected reasoning across perception, prediction, planning, and behavior-level driving decisions.
OmniDrive~\cite{OmniDrive} introduces a VQA dataset focusing on counterfactual reasoning, providing multi-view images, trajectory simulations, and decision-making QA pairs. 
WOMD-Reasoning~\cite{WOMD-Reasoning}, built on the Waymo Open Motion Dataset, provides camera image embeddings and LiDAR data alongside object trajectories focusing on multi-agent interaction reasoning and traffic rule-induced behaviors. 
%
%

Although prior work has established LLMs as viable components for autonomous driving tasks, CommandLM introduces distinct technical contributions addressing complementary objectives. While CommandLM reuses BLIP-2's Q-Former architecture, it extends it with temporal BEV fusion to reason over scene evolution rather than isolated frames. Unlike EMMA and OmniDrive, which operate with limited temporal context or construct intermediate world models, CommandLM employs explicit temporal context encoding through a rolling buffer of stacked BEV feature maps, enabling continuous reasoning over scene evolution. Furthermore, CommandLM frames the problem as behavior-level captioning with object-centric semantic grounding, in contrast to LMDrive's formulation as end-to-end control. This differs from DriveLM's Graph VQA approach, which relies on coordinate-based spatial references. The CommandLM dataset augmentation strategy prioritizes linguistically diverse QA pairs grounded in natural language object references, replacing coordinate-heavy formulations and introducing additional QA samples focused on behavioral exploration. Together, these design choices position CommandLM as a framework for behavior-level interpretation of autonomous driving scenarios through multimodal fusion and temporal reasoning.


\section{Method}
\subsection{Architecture}
\begin{figure}[h]
    \centering
    \resizebox{0.47\textwidth}{!}{%
\definecolor{visionblue}{RGB}{65, 105, 225}
\definecolor{nlpgreen}{RGB}{34, 139, 34}
\definecolor{processgray}{RGB}{70, 70, 70}
\definecolor{lightgray}{RGB}{245, 245, 245}
\definecolor{accentorange}{RGB}{255, 140, 0}
\definecolor{tokenvision}{RGB}{100, 149, 237}
\definecolor{tokentext}{RGB}{60, 179, 113}
\definecolor{outputred}{RGB}{220, 53, 69}

\begin{tikzpicture}[
        scale=0.9,
        font=\sffamily,
        >=latex,
        box/.style={
                draw=black!60,
                fill=lightgray,
                rounded corners=3pt,
                drop shadow={opacity=0.15, shadow xshift=1pt, shadow yshift=-1pt}
            },
        visionbox/.style={
                draw=visionblue!80,
                fill=visionblue!10,
                rounded corners=3pt,
                drop shadow={opacity=0.15, shadow xshift=1pt, shadow yshift=-1pt}
            },
        nlpbox/.style={
                draw=nlpgreen!80,
                fill=nlpgreen!10,
                rounded corners=3pt,
                drop shadow={opacity=0.15, shadow xshift=1pt, shadow yshift=-1pt}
            },
        processbox/.style={
                draw=processgray!80,
                fill=processgray!8,
                rounded corners=3pt,
                drop shadow={opacity=0.15, shadow xshift=1pt, shadow yshift=-1pt}
            },
        llmbox/.style={
                draw=accentorange!80,
                fill=accentorange!15,
                rounded corners=3pt,
                drop shadow={opacity=0.25, shadow xshift=2pt, shadow yshift=-2pt},
                line width=1.5pt
            },
        arrow/.style={
                ->,
                thick,
                color=processgray!80,
                line width=2pt
            },
        bigarrow/.style={
                ->,
                very thick,
                color=accentorange!80,
                line width=3pt
            }
    ]

    \def\boxwidth{6.5}
    \def\boxheight{3}
    \def\midline{1.5}
    \def\vspacing{1.5}

    \def\smallboxwidth{3}
    \def\smallboxheight{2.5}

    \draw[visionbox] (0,0) rectangle (\smallboxwidth,\smallboxheight);
    \node[text width=2.8cm, align=center, font=\large\bfseries, color=visionblue!80]
    at (\smallboxwidth/2,\smallboxheight*0.7) {LIDAR};
    \node[text width=2.8cm, align=center, font=\small\bfseries]
    at (\smallboxwidth/2,\smallboxheight*0.3) {Point Cloud}; 

    \draw[visionbox] (\smallboxwidth + 0.5,0) rectangle (2*\smallboxwidth + 0.5,\smallboxheight);
    \node[text width=2.8cm, align=center, font=\large\bfseries, color=visionblue!80]
    at (\smallboxwidth + 0.5 + \smallboxwidth/2,\smallboxheight*0.7) {CAMERA};
    \node[text width=2.8cm, align=center, font=\small\bfseries]
    at (\smallboxwidth + 0.5 + \smallboxwidth/2,\smallboxheight*0.3) {Multi-view RGB}; 

    \draw[nlpbox] (\boxwidth + 2.6,0) rectangle (\boxwidth + 2.6 + 5.2,2.4);

    \node[text width=4.8cm, align=center, font=\large\bfseries, color=nlpgreen!80]
    at (\boxwidth + 2.6 + 2.6,2.4*0.7) {QUESTION};
    \node[text width=4.8cm, align=center, font=\small\bfseries]
    at (\boxwidth + 2.6 + 2.6,2.4*0.3) {What is the vehicle doing?};

    \def\middleboxheight{2.7}
    \def\middlemidline{1.4}
    \def\middleboxy{-\middleboxheight - \vspacing}

    \draw[processbox] (0,\middleboxy) rectangle (\boxwidth,\middleboxy + \middleboxheight);
    \draw[processgray!40] (0,\middleboxy + \middlemidline) -- (\boxwidth,\middleboxy + \middlemidline);

    \node[text width=5.5cm, align=center, font=\bfseries]
    at (\boxwidth/2,\middleboxy + \middlemidline + \middlemidline/2 +0.1) {BEVFusion}; 
    \node[text width=5.5cm, align=center, font=\footnotesize, color=processgray] at (\boxwidth/2,\middleboxy + \middlemidline + \middlemidline/2 - 0.4) {$512 \times 32400$};
    \node[text width=5.5cm, align=center, font=\bfseries]
    at (\boxwidth/2,\middleboxy + \middlemidline/2 + 0.3) {BUFFER};
    \node[text width=5.5cm, align=center, font=\footnotesize\itshape, color=processgray]
    at (\boxwidth/2,\middleboxy + \middlemidline/2 - 0.2) {up to 2 seconds stacked feature maps};
    \node[text width=5.5cm, align=center, font=\footnotesize, color=processgray] at (\boxwidth/2,\middleboxy + \middlemidline/2 - 0.5) {$24 \times 512 \times 32400$};

    \draw[arrow] (\smallboxwidth/2,0) -- (\smallboxwidth/2,\middleboxy + \middleboxheight);
    \draw[arrow] (\smallboxwidth + 0.5 + \smallboxwidth/2,0) -- (\smallboxwidth + 0.5 + \smallboxwidth/2,\middleboxy + \middleboxheight);

    \def\bottomboxheight{3.2}
    \def\bottomsegheight{1.067}
    \def\bottomspacing{2.5}
    \def\bottomboxy{\middleboxy - \middleboxheight - \bottomspacing}

    \draw[processbox] (0,\bottomboxy) rectangle (\boxwidth,\bottomboxy + \bottomboxheight);
    \draw[processgray!40] (0,\bottomboxy + \bottomsegheight) -- (\boxwidth,\bottomboxy + \bottomsegheight);
    \draw[processgray!40] (0,\bottomboxy + 2*\bottomsegheight) -- (\boxwidth,\bottomboxy + 2*\bottomsegheight);

    \node[text width=5.5cm, align=center, font=\bfseries]
    at (\boxwidth/2,\bottomboxy + 2.8) {Adapter Down};
    \node[text width=5.5cm, align=center, font=\footnotesize, color=processgray] at (\boxwidth/2,\bottomboxy + 2.4) {$12288 \times 1408$};
    \node[text width=5.5cm, align=center, font=\bfseries]
    at (\boxwidth/2,\bottomboxy + 1.8) {Q-Former};
    \node[text width=5.5cm, align=center, font=\footnotesize, color=processgray] at (\boxwidth/2,\bottomboxy + 1.3) {$32 \times 768$};
    \node[text width=5.5cm, align=center, font=\bfseries]
    at (\boxwidth/2,\bottomboxy + 0.6) {Adapter Up};
    \node[text width=5.5cm, align=center, font=\footnotesize, color=processgray] at (\boxwidth/2,\bottomboxy + 0.25) {$32 \times 5120$};

    \draw[arrow] (\boxwidth/2,\middleboxy) -- (\boxwidth/2,\bottomboxy + \bottomboxheight);

    \def\tokenizerheight{1.8}
    \def\tokenizerboxy{-5.35}

    \draw[processbox] (\boxwidth + 2,\tokenizerboxy) rectangle (2*\boxwidth + 2,\tokenizerboxy + \tokenizerheight);

    \node[text width=5.5cm, align=center, font=\large\bfseries, color=processgray!80]
    at (\boxwidth + 2 + \boxwidth/2,\tokenizerboxy + \tokenizerheight/2) {TOKENIZER};

    \draw[arrow] (\boxwidth + 2 + \boxwidth/2,0) -- (\boxwidth + 2 + \boxwidth/2,\tokenizerboxy + \tokenizerheight);

    \def\squaresize{0.6}
    \def\squarey{-10.75}
    \def\tokengap{0.65}

    \foreach \i in {0,1,2,3,4,5,6,7,8,9} {
            \fill[tokenvision, rounded corners=2pt, drop shadow={opacity=0.2}]
            (\i*\tokengap,\squarey) rectangle (\i*\tokengap + \squaresize,\squarey + \squaresize);
        }

    \foreach \i in {0,1,2,3,4,5,6,7,8,9} {
            \fill[tokentext, rounded corners=2pt, drop shadow={opacity=0.2}]
            (8.5 + \i*\tokengap,\squarey) rectangle (8.5 + \i*\tokengap + \squaresize,\squarey + \squaresize);
        }

    \draw[arrow] (\boxwidth/2,\bottomboxy) -- (\boxwidth/2,\squarey + \squaresize);
    \draw[arrow] (\boxwidth + 2 + \boxwidth/2,\tokenizerboxy) -- (\boxwidth + 2 + \boxwidth/2,\squarey + \squaresize);

    \draw[llmbox] (0.5,-13.5) rectangle (14.5,-11.5);
    \draw[accentorange!60, line width=1.5pt] (0.5,-12.5) -- (14.5,-12.5);

    \node[font=\LARGE\bfseries, color=accentorange!80] at (7.5,-12.0) {LARGE LANGUAGE MODEL};
    \node[text width=13cm, align=center, font=\small] at (7.5,-13.0)
    {Qwen 2.5 $\bullet$ 14B parameters $\bullet$ 8-bit quantized $\bullet$ LoRA fine-tuned};

    \draw[bigarrow] (3.2,\squarey) -- (3.2,-11.5);
    \draw[bigarrow] (11.8,\squarey) -- (11.8,-11.5);

    \def\newboxheight{2}
    \def\newboxy{-16.5}

    \draw[nlpbox] (3.5,\newboxy) rectangle (11.5,\newboxy + \newboxheight);
    \draw[nlpgreen!60] (3.5,\newboxy + \newboxheight/2) -- (11.5,\newboxy + \newboxheight/2);

    \node[text width=7cm, align=center, font=\large\bfseries, color=nlpgreen!80]
    at (7.5,\newboxy + \newboxheight*0.75) {Behavior Level Caption};
    \node[text width=7cm, align=center, font=\small\bfseries]
    at (7.5,\newboxy + \newboxheight*0.25) {Braking as the traffic light turns red.};

    \draw[bigarrow] (7.5,-13.5) -- (7.5,\newboxy + \newboxheight);

\end{tikzpicture}
    }
    \caption{CommandLM architecture: Temporal BEV fusion → Q-Former compression → LLM behavior generation. Blue squares: visual tokens; green squares: text tokens.}
    \label{fig:cmdlm}
\end{figure}
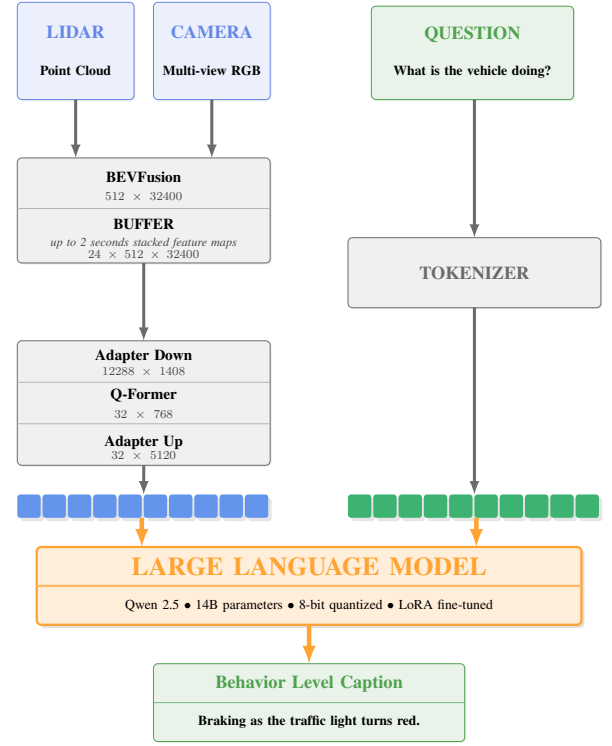

\noindent
CommandLM retains the three-stage multimodal stack of Vision Encoder, Adapter, and LLM, leveraging components such as BEVFusion~\cite{BEVFusion}, Q-Former~\cite{BLIP2}, and LLM.  A key innovation is the spatio-temporal compression bottleneck, where multi-frame BEV dynamics are distilled into a dense query representation, enabling reasoning over scene evolution. An overview of CommandLM is presented in Fig.~\ref{fig:cmdlm}. Below, we describe the architecture in detail, structured into its core modules.

\subsubsection{Multimodal Sensor Fusion via BEV}
CommandLM employs BEVFusion to fuse six RGB cameras and LiDAR, providing each frame with both detailed semantic richness from images and precise geometric structure from the point cloud. Camera images are processed through a Swin Transformer backbone and projected to BEV via depth estimation, while LiDAR undergoes voxelization and projects directly. Both modalities are concatenated to capture the dynamic structure of driving scenes.

\subsubsection{Temporal Context Encoding}
CommandLM’s ability to reason like a human relies on temporal context. Instead of processing only the current frame, it uses a rolling buffer of BEV tensors that typically cover two seconds at 12 Hz and are stacked and reshaped for downstream processing. This enables the model to capture not just static layouts but also scene evolution, such as lane changes, vehicle approaches, and pedestrian movement. By maintaining this short-term memory, similar to how a driver tracks surrounding motion, CommandLM achieves more continuous perception and better anticipates dynamic interactions.

\subsubsection{Compression via Q-Former}
To bridge the gap between high-dimensional visual features and the language domain, CommandLM employs the Q-Former module, adapted from BLIP-2. The Q-Former serves as a learned bottleneck and acts as an adapter between vision and language, compressing the temporal BEV sequence into a set of 32 query tokens. Each query token is a learned representation of the most relevant scene elements. These queries interact with the BEV sequence via cross-attention layers to inform downstream behavioral reasoning. By structuring temporal knowledge in stacked frames before passing it to the Q-Former, the model abstracts over time, not just within a single image.

\subsubsection{Vision-Language Interface and LLM Integration}
After compression, a second adapter bridges the Q-Former outputs to the LLM backbone, creating vision tokens that ensure seamless integration with the LLM. To preserve modality boundaries, the vision tokens are enclosed with special delimiter tokens: \texttt{<vision\_start>} and \texttt{<vision\_end>}. These delimiters signal to the LLM that the upcoming segment contains multimodal content and should be interpreted as visual prefix to its own text for downstream reasoning. Finally, this setup allows the LLM to perform perception, prediction, planning, or behavior tasks, conditioned on rich visual-temporal context, within its autoregressive generation process.

\subsection{Training details}
\noindent
For the language modeling component, CommandLM employs Qwen-2.5 14B, a decoder-only LLM chosen for its strong performance among LLMs below 32 billion parameters and its flexibility (including for hardware-limited training and inference). To facilitate both rapid experimentation and resource efficiency, only the two adapter layers, the full Q-Former, and LoRA parameters for the LLM are updated, while BEVFusion remains frozen. By using 8-bit quantization on the LLM, we ensure that CommandLM can maintain fast inference speeds (average 1.78 seconds) on a user-grade GPU without the compromise in generation quality typically observed in smaller models. 
Training is orchestrated using the AdamW optimizer with a schedule-free regime, foregoing explicit learning rate schedules. This approach eliminates manual tuning and enables aggressive, yet stable convergence. Learning rates are module-specific: Adapters train at $5 \times 10^{-4}$ since they are newly initialized, while the Q-Former and LoRA matrices train at $1 \times 10^{-4}$. Momentum and decay are set at $\beta_1 = 0.9$, $\beta_2 = 0.999$, and a weight decay of 0.05, aligning with community standards for fine-tuning large networks. LoRA is configured with rank 16 and alpha 32, with a dropout rate of 0.05 to prevent overfitting. The model is trained on batches of 10 scenes with gradient accumulation across 5 NVIDIA A6000 GPUs using 8-bit quantization combined with mixed-precision training. The system is trained over a single epoch due to resource constraints, converging within four days.

\section{Experiments}
\subsection{Dataset augmentation}
\noindent
The CommandLM-nuScenes dataset is derived from the DriveLM-nuScenes~\cite{DriveLM} subset, a structured resource designed for VQA within autonomous driving scenarios. DriveLM provides extensive graph-structured QA pairs. Each graph node corresponding to a scene-specific question (spanning perception, prediction, planning, and driving behavior), with logical edges reflecting dependencies typical of human reasoning. While this structure offers a strong starting point for behavior-level language grounding, DriveLM’s original format is not directly suited to training this works model due to two primary issues: (1) an overrepresentation of pixel- or coordinate-based references that tend to confuse the model, and (2) relatively limited linguistic diversity in both questions and answers.
To better align the dataset with CommandLM’s intended output, which is a concise, context-rich caption describing the ego vehicle’s next maneuver in naturalistic language, an extensive augmentation process was implemented. All VQA samples from the DriveLM-nuScenes subset were first sanitized. Rather than encoding objects with explicit pixel coordinates and camera-specific information (e.g., "at pixel (162, 732) in the right camera"), we replaced these with abstract object references using special tags (e.g. \texttt{Q: What is the visual description of <|object\_ref\_start|>obj 2<|object\_ref\_end|>? A: Black SUV}). This shift towards object references in the BEV space, allows the model to reason naturally about semantic entities and their spatial relationships without requiring pixel decoding, which LLMs struggle with due to their inherent numerical insensitivity.~\cite{Wen2025, chen-etal-2023-improving}.
In the next stage, the dataset was vastly expanded to improve both data volume and linguistic diversity through a structured two-stage pipeline using Qwen-2.5 (72 billion parameters). 
The pipeline leveraged the original QA pairs, comprising approximately 100–150 per annotated frame across three task categories (perception, prediction and planning) and a single QA for the behavior category.
%
First, the model extracts comprehensive scene summaries by synthesizing the task-specific QA responses from all three categories (excluding behavior). These summaries capture the ego vehicle's kinematic state, surrounding dynamic agents, static infrastructure, and prioritize hazards, expressed in location-neutral language without absolute coordinates. Thus establishing a textual scaffold for data augmentation.
Next, leveraging the structured scene summary, the model was prompted to generate ten behavior-focused questions per frame, ranging from basic operational directives ("What is the safest immediate behavior for the ego vehicle?") to strategic planning queries ("How should the vehicle plan its maneuvers given the current mix of dynamic road users?"). 
Finally, answers were generated by the same model using the newly created questions, grounding each response exclusively in information recoverable from the structured scene summary. This augmentation process generated over 40{,}000 additional QA pairs, allowing CommandLM to systematically explore the space of possible vehicle behaviors.

\subsection{Experimental Setup}
\noindent
The experimental evaluation of CommandLM is grounded in a framework, designed to measure both the linguistic and contextual quality of captions generated. All experiments are conducted on the CommandLM-nuScenes dataset. Each sample consists of a stacked and fused BEV tensor representing up to two seconds of driving context at 12\,Hz, paired with natural language questions, purposefully stripped of coordinate-heavy details to focus the model on object-level semantics and scene dynamics.

\subsection{Quantitative Results}
\noindent
To benchmark the trained model's performance in generating behavior-level driving captions, the evaluation metrics summarized in Table~\ref{tab:evaluation-metrics} are chosen for their ability to assess complementary aspects of caption quality and thus provide a comprehensive view of text-generation performance.
Since existing MLLMs for ego vehicles either lack publicly available weights or are not addressing behavior-level captioning, we train BLIP-2 on the same augmented dataset to establish a reference point for evaluating CommandLM's performance. 
BLIP-2 and CommandLM share a common foundation in the Q-Former module, but CommandLM distinguishes itself by incorporating temporal BEV fusion, which is absent in BLIP-2. This architectural difference provides a clear basis for isolating and evaluating the specific contributions of temporal fusion to model performance.
\begin{table}[h]
        \centering
        \begin{tabular}{|l|c|c|}
            \hline
            \textbf{Metric $\uparrow$} & \textbf{BLIP-2} & \textbf{CommandLM} \\
            \hline
            METEOR                     & 0.1023          & 0.3822             \\
            CIDEr                      & 0.0710          & 0.6687             \\
            SPICE                      & 0.0738          & 0.1337             \\
            BERT-P                     & 0.8394          & 0.8548             \\
            BERT-R                     & 0.8797          & 0.8974             \\
            BERT-F1                    & 0.8575          & 0.8756             \\
            \hline
        \end{tabular}
        \vspace{1mm}
        \caption{METEOR measures fluency and grammatical correctness, CIDEr evaluates distinctive n-grams reflecting human consensus, SPICE assesses semantic grounding via scene graph matching, and BERTScore compares contextual embeddings to capture semantic similarity. Higher values indicate stronger linguistic quality, semantic relevance, and scene alignment.}
        \label{tab:evaluation-metrics}
\end{table} 
\noindent
The metrics in Table~\ref{tab:evaluation-metrics} evaluate our model’s descriptive performance using complementary language generation measures.
METEOR assesses fluency and grammatical correctness based on word- and phrase-level matches.
CIDEr measures the use of semantically salient and distinctive n-grams that align with human consensus, rewarding rare but informative phrases that capture critical scene details.
BERTScore complements this by comparing contextual embeddings from pre-trained transformers, allowing recognition of semantic equivalence and paraphrases beyond surface wording.
SPICE focuses on semantic grounding by decomposing captions into scene graphs of objects, attributes, and relations to assess whether the generated descriptions reflect the actual visual content.
CommandLM achieves higher scores across all reported metrics compared to BLIP-2. Specifically, CommandLM scores 0.67 CIDEr versus 0.52 for BLIP-2, 0.88 BERT-F1 versus 0.73, and 0.38 METEOR versus 0.10. These quantitative differences suggest advantages across linguistic quality, semantic relevance, and grounding in scene understanding.
%
%
%
%
\begin{figure}[h]
        \begin{tikzpicture}
            \begin{axis}[
                ybar,
                bar width=12pt,
                width=0.5\textwidth,
                height=0.25\textwidth,
                ylabel={Percentage (\%)},
                xlabel={Driving Action},
                symbolic x coords={Left, Right, Straight, Stop},
                xtick=data,
                ymin=0,
                ymax=65,
                ymajorgrids=true,
                grid style=dashed,
                enlarge x limits=0.15,
            ]
            \addplot[fill=blue!60, draw=blue!80!black] coordinates {
                (Left, 17.29)
                (Right, 22.97)
                (Straight, 28.45)
                (Stop, 31.29)
            };
            \addplot[fill=red!60, draw=red!80!black] coordinates {
                (Left, 12.50)
                (Right, 21.57)
                (Straight, 55.15)
                (Stop, 10.78)
            };
            \addplot[fill=green!60, draw=green!80!black] coordinates {
                (Left, 9.56)
                (Right, 11.27)
                (Straight, 58.34)
                (Stop, 20.83)
            };
        \end{axis}
        \end{tikzpicture}
        \caption{Distribution of driving behaviors predicted by BLIP-2 (blue) and CommandLM (red), compared against the ground-truth dataset distribution (green). The plot summarizes the frequency of four behavior classes (left turn, right turn, straight, and stop) showing how closely each model’s predictions align with real driving patterns.}
        \label{fig:mllm_results_bar}
\end{figure}
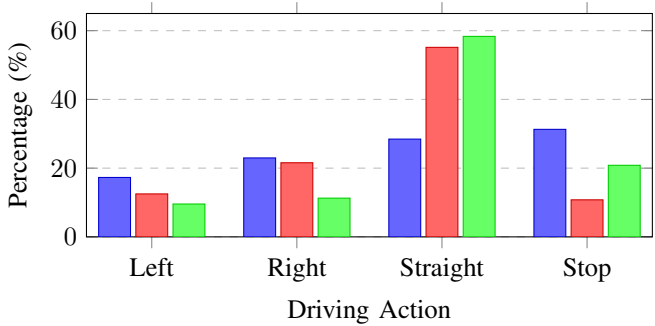
\noindent
Beyond standard metrics, we conducted a custom evaluation, shown in Fig.~\ref{fig:mllm_results_bar}, in which the model was prompted to identify the most likely behavior for each sample, yielding a distribution of predicted actions such as going straight, turning left or right, and stopping. This approach enables us to assess both the model’s confidence and its consistency across comparable scenarios. The resulting distributions align with typical urban driving patterns, where going straight occurs most frequently, followed by turns and stops. Notably, CommandLM demonstrates superior alignment with the dataset distribution compared to BLIP-2, suggesting that it more accurately captures human driving behavior patterns observed in real-world scenarios.
\begin{table}[t]
    \begin{minipage}{0.5\textwidth}
        \centering
        \begin{tabular}{|l|c|c|}
            \hline
                                            & \textbf{BLIP-2} & \textbf{CommandLM} \\
            \hline
            \textbf{Behavior Exploration Rate} & 41.61\%         & 58.00\%            \\
            \textbf{Ground Truth Matching}     & 34.52\%         & 47.06\%            \\
            \hline
        \end{tabular}
        \vspace{1mm}
        \caption{Comparison on two human-validated behavior metrics. The Behavior Exploration Rate is the percentage of predicted actions judged feasible within the driving scene. Ground Truth Matching is the percentage of predictions that correspond to the actual human driving behavior. Higher values indicate stronger behavioral plausibility and accuracy.}
        \label{tab:mllm_acc_results}
    \end{minipage}
\end{table}
\noindent
To assess CommandLM's ability to generate plausible and accurate driving behaviors, we evaluate two complementary metrics through human-in-the-loop validation in Table~\ref{tab:mllm_acc_results}. For each test sample, the model is prompted to predict the ego vehicle's behavior. These predictions are compared against map topology constraints and ground truth annotations.
The Behavior Exploration Rate measures the model's capacity to generate valid, rule-compliant driving behaviors that extend beyond the original ground truth annotations. When the model generates a behavior caption a human validator manually assesses whether the predicted action is feasible given the map topology and traffic rules for that sample. However, behaviors that are technically valid but inefficient or non-goal-oriented, such as stopping during continuous cruising segments, detract from model performance. This metric captures the model's ability to recognize multiple valid behavioral options for a given driving context, with CommandLM achieving 58\% compared to BLIP-2's 41.61\%.
The Ground Truth Matching metric evaluates how accurately the model reproduces the actual human driving behavior observed in the dataset. For each sample, the model generates a behavior caption, and if this caption semantically aligns with the ground truth driving decision, it is marked as correct, indicating that the model successfully predicted how a human driver would act in that situation. Deviations from the ground truth are considered incorrect, as they represent cases where the model fails to match real-world human driving patterns. CommandLM demonstrates superior performance at 47.06\% versus BLIP-2's 34.52\%, indicating moderate yet promising alignment with real-world driving expectations.

\subsection{Qualitative Results}
\noindent
The qualitative assessment of CommandLM’s output provides deeper insight into its real-world utility beyond what automated metrics can measure. 
\begin{figure}[h]
    \centering
    \includegraphics[width=0.47\textwidth]{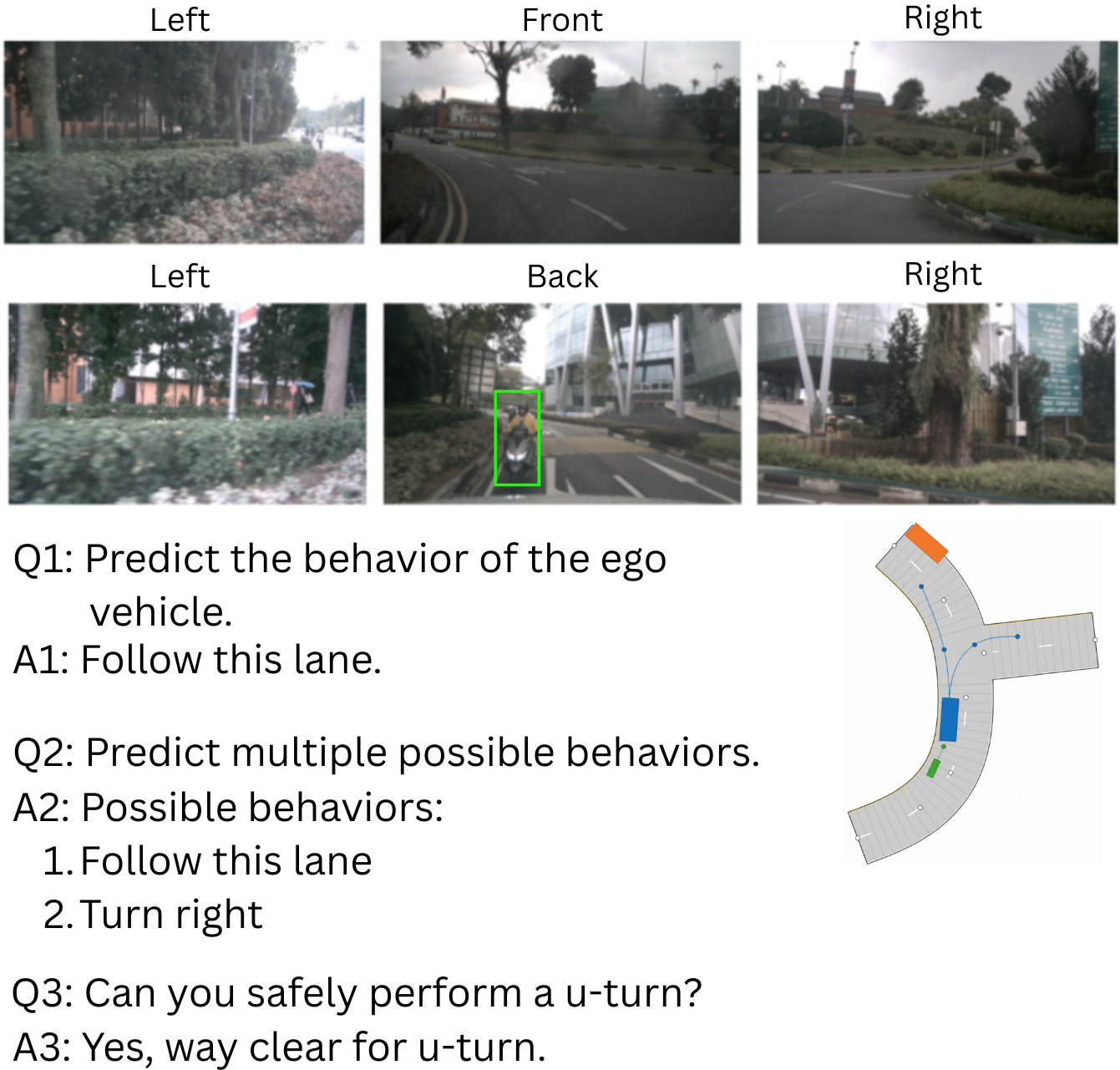}
    \caption{Behavior prediction from LiDAR and multi-view cameras. Top-down view shows ego (blue), trailing motorcycle (green), and stationary vehicle (orange), with multiple predicted ego trajectories illustrating context-aware reasoning and anticipated actions.}
    \label{fig:uturn-example}
\end{figure}
Manual inspection of generated captions from the nuScenes test set reveal the emergence of human-like reasoning and situational abstraction, as illustrated in Fig.~\ref{fig:uturn-example}.
The model demonstrates versatility from basic behavior prediction ("follow this lane") to ranking them ("possible behaviors: straight, right") and context-aware maneuver assessment ("yes, way clear for u-turn"), showcasing adaptive linguistic complexity and scene understanding.
CommandLM adapts its directives based on evolving situation complexity as seen in Fig.~\ref{fig:pedestrian}, progressing from routine navigation ("follow the curve to the right") to conflict avoidance ("yield to the crossing pedestrian"), and ultimately to defensive maneuvering ("slow down and prepare to stop"). This graduated response strategy demonstrates temporal reasoning and behavioral anticipation.
In human evaluations, CommandLM’s captions are rated as plausible and actionable, frequently matching the rationale and tone preferred by domain experts. In several examples, the model not only captures literal scene content but constructs coherent explanations useful for downstream planning like, rerouting around hazards or explaining regulatory stops.
Overall, both quantitative and qualitative evaluations confirm that CommandLM advances the state of behavior-level comprehension in real-world, multimodal autonomous driving settings, generating not just accurate but interpretable and deployable driving scene captions.

\section{Discussion}
\noindent
CommandLM demonstrates that MLLMs can generate interpretable, behavior-level driving captions from sensor data using limited computational resources. Our evaluation shows that the proposed model outperforms the baseline across several key metrics, confirming that this approach constitutes meaningful progress.
Despite these promising results in Table~\ref{tab:evaluation-metrics}, the relatively low SPICE score suggests that the model occasionally hallucinates or produces less semantically faithful descriptions.
To obtain a more detailed understanding of model performance, the metrics reported in Table~\ref{tab:mllm_acc_results} indicate that the model can explore diverse possible behaviors rather than merely reproducing a fixed pattern. However, the moderate performance in following the human ground-truth path suggests that, while behavior-level caption generation is viable, achieving precise path fidelity remains a challenge.
Several factors may contribute to these observations. First, the compression applied by the Q-Former module may reduce high-dimensional feature maps too aggressively, potentially discarding important information relevant to driving scenarios.
%
To address this, an attention-based memory bank could be incorporated to propagate information from previous vision tokens to subsequent ones, thereby conveying historical context to the current frame~\cite{OmniDrive, MA-LMM}. The goal is to maintain a compact yet contextually rich set of vision tokens, enabling the model to focus on the most relevant features without being overwhelmed by redundant context.
Second, the underlying nuScenes dataset may not represent edge cases to a sufficient degree. While nuScenes provides a rich multimodal corpus of urban traffic scenes, it underrepresents rare or hazardous events, such as near-collisions or extreme weather, which limits the model’s exposure to atypical driving conditions.
Adding the Waymo Open Dataset~\cite{WaymoOpenDataset}, with its diverse scenarios to our dataset might create a more varied mix of real-world scenarios. Specific underrepresented edge cases can be simulated with CARLA, focusing on situations that are difficult to capture at scale in real-world datasets.
%
Following the methodology of DriveLM~\cite{DriveLM}, a hybrid annotation approach could be employed, using a rule-based system for initial labeling. To enrich this with linguistic diversity we could add a reformulation step with a LLM similarly to our approach. Training on such a combined dataset would expose CommandLM to the distributional variety necessary for robust behavior captioning across both routine and safety-critical driving situations.

\section{Conclusion}
\noindent
Our work presented CommandLM, a MLLM designed to generate interpretable, behavior-level driving captions from fused sensor data. The proposed framework demonstrates that behavior reasoning can be achieved efficiently through temporal BEV fusion and language-based abstraction, without sacrificing interpretability or computational feasibility. 
Beyond quantitative improvements over the BLIP-2 baseline, the primary contribution of this work lies in establishing behavior-level captioning as a viable interface between perception and decision-making. By translating complex scene dynamics into human-readable descriptions, CommandLM supports transparent validation, planner supervision, and safer human–machine collaboration.
Future work will focus on tighter coupling between caption generation and downstream control, extending temporal reasoning to multi-agent interactions.


\bibliographystyle{IEEEtran}
\bibliography{references}

@inproceedings{BEVFusion,
  title={BEVFusion: Multi-Task Multi-Sensor Fusion with Unified Bird's-Eye View Representation},
  author={Liu, Zhijian and Tang, Haotian and Amini, Alexander and Yang, Xingyu and Mao, Huizi and Rus, Daniela and Han, Song},
  booktitle={Proceedings of the IEEE International Conference on Robotics and Automation},
  year={2023}
}

@inproceedings{ViTs,
  title={An Image is Worth 16x16 Words: Transformers for Image Recognition at Scale},
  author={Anton Dosovitskiy and Luca Beyer and Alexander Kolesnikov and Dirk Weissenborn and Xiaohua Zhai and Thomas Unterthiner and Mostafa Dehghani and Matthias Minderer and Georg Heigold and Sylvain Gelly and Jakob Uszkoreit and Neil Houlsby},
  booktitle={Proceedings of the International Conference on Learning Representations},
  year={2021}
}

@inproceedings{BLIP2,
  title={BLIP-2: Bootstrapping Language-Image Pre-training with Frozen Image Encoders and Large Language Models},
  author={Junnan Li and Dongxu Li and Silvio Savarese and Steven Hoi},
  booktitle={Proceedings of the 40th International Conference on Machine Learning},
  year={2023},
}

@inproceedings{DrivingWithLLM,
  title={Driving with LLMs: Fusing Object-Level Vector Modality for Explainable Autonomous Driving},
  author={Chen, Long and Sinavski, Oleg and Hünermann, Jan and Karnsund, Alice and Willmott, Andrew James and Birch, Danny and Maund, Daniel and Shotton, Jamie},
  booktitle={Proceedings of the IEEE International Conference on Robotics and Automation},
  year={2024}
}

@article{Atakishiyev2021,
  title={Explainable Artificial Intelligence for Autonomous Driving: A Comprehensive Overview and Field Guide for Future Research Directions},
  author={Atakishiyev, Shahin and Salameh, Mohammad and Yao, Hengshuai and Goebel, Randy},
  journal={IEEE Access},
  year={2021},
}

@article{Chen2024,
  title={End-to-End Autonomous Driving: Challenges and Frontiers},
  author={Chen, Xiaosong and Zhang, Tuo and Wang, Yuhui and Wang, Song and Hebert, Martial},
  journal={IEEE Transactions on Pattern Analysis and Machine Intelligence},
  year={2024}
}

@inproceedings{Wen2025,
  title={LeAD: The LLM Enhanced Planning System Converged End-to-End Driver},
  author={Wen, Hao and Zhong, Jing and Wu, Yanbo and Qi, Ruoying and Sun, Boyang and Yao, Yuanzhe and Gu, Jiaqi},
  booktitle={Proceedings of the International Conference on Learning Representations},
  year={2025}
}

@article{Acharya2025,
  author = {Acharya, Rajani},
  title = {LLM integration in autonomous vehicle systems},
  journal = {World Journal of Advanced Research and Reviews},
  year = {2025},
}

@article{Cui2025Survey,
  title = {A Survey on Large Language Model-Powered Autonomous Driving},
  journal = {ScienceDirect Engineering},
  year = {2025},
  author = {Yuxuan, Zhu and Shiyi, Wang and Wenqing, Zhong and Nianchen, Shen and Yunqi, Li and Siqi, Wang and Zhiheng, Li and Cathy, Wu and Zhengbing, He and et al.}
}

@inproceedings{chen-etal-2023-improving,
    title = "Improving Numeracy by Input Reframing and Quantitative Pre-Finetuning Task",
    author = "Chen, Chung-Chi  and
      Takamura, Hiroya  and
      Kobayashi, Ichiro  and
      Miyao, Yusuke",
    booktitle = "Findings of the European Chapter
    of the Association for Computational Linguistics",
    year = "2023",

}

@article{EMMA,
  title={EMMA: End-to-End Multimodal Model for Autonomous Driving},
  author={Jyh-Jing Hwang and Runsheng Xu and Hubert Lin and Wei-Chih Hung and Jingwei Ji and Kristy Choi and Di Huang and Tong He and Paul Covington and Benjamin Sapp and et al.},
  journal={Transactions on Machine Learning Research},
  year={2025},
}

@inproceedings{AsyncDriver,
  author = {Yuan Chen and Zi-han Ding and Ziqin Wang and Yan Wang and Lijun Zhang and Si Liu},
  title = {Asynchronous Large Language Model Enhanced Planner for Autonomous Driving},
  booktitle={Proceedings of the European Conference on Computer Vision},
  year = {2024}
}

@inproceedings{LeapAD,
  title={Continuously Learning, Adapting, and Improving: A Dual-Process Approach to Autonomous Driving},
  author={Jianbiao Mei and Yukai Ma and Xuemeng Yang and Licheng Wen and Xinyu Cai and Xin Li and Daocheng Fu and Bo Zhang and Pinlong Cai and Min Dou and et al.},
  booktitle={Proceedings of the Advances in Neural Information Processing Systems},
  year={2024}
}

@inproceedings{MotionLM,
  title={MotionLM: Multi-Agent Motion Forecasting as Language Modeling},
  author={Ari Seff and Brian Cera and Dian Chen and Mason Ng and Aurick Zhou and Nigamaa Nayakanti and Khaled S. Refaat and Rami Al-Rfou and Benjamin Sapp},
  booktitle={Proceedings of the IEEE/CVF International Conference on Computer Vision},
  year={2023}
}

@inproceedings{VisionTrap,
  title={Visiontrap: Vision-augmented trajectory prediction guided by textual descriptions},
  author={Moon, Seokha and Woo, Hyun and Park, Hongbeen and Jung, Haeji and Mahjourian, Reza and Chi, Hyung-gun and Lim, Hyerin and Kim, Sangpil and Kim, Jinkyu},
  booktitle={Proceedings of the European Conference on Computer Vision},
  year={2024},
}

@inproceedings{DriveLM,
  title={DriveLM: Driving with Graph Visual Question Answering},
  author={Chonghao Sima and Katrin Renz and Kashyap Chitta and Li Chen and Hanxue Zhang and Chengen Xie and Ping Luo and Andreas Geiger and Hongyang Li},
  booktitle={Proceedings of the European Conference on Computer Vision},
  year={2024}
}

@inproceedings{WaymoOpenDataset,
  author={Sun Pei and Kretzschmar Henrik and Dotiwalla Xerxes and Chouard Aurélien and Patnaik Vijaysai and Tsui Paul and Guo James and Zhou Yin and Chai Yuning and Caine Benjamin and et al.},
  booktitle={Proceedings of the IEEE/CVF Conference on Computer Vision and Pattern Recognition}, 
  title={Scalability in Perception for Autonomous Driving: Waymo Open Dataset}, 
  year={2020}
}

@article{BDD-X,
  title={Textual Explanations for Self-Driving Vehicles},
  author={Kim Jinkyu and Rohrbach Anna and Darrell Trevor and Canny, John and Akata, Zeynep},
  journal={Proceedings of the European Conference on Computer Vision},
  year={2018}
}

@inproceedings{WOMD-Reasoning,
  title={WOMD-Reasoning: A Large-Scale Dataset and Benchmark for Interaction and Intention Reasoning in Driving},
  author={Yiheng Li and Cunxin Fan and Chongjian GE and Seth Zhao and Chenran Li and Chenfeng Xu and Huaxiu Yao and Masayoshi Tomizuka and Bolei Zhou and Chen Tang and et al.},
  booktitle={Proceedings of the 42nd International Conference on Machine Learning},
  year={2025}
}

@inproceedings{DriveVLM,
  title={DriveVLM: The Convergence of Autonomous Driving and Large Vision-Language Models},
  author={Xiaoyu Tian and Junru Gu and Bailin Li and Yicheng Liu and Yang Wang and Zhiyong Zhao and Kun Zhan and Peng Jia and Xianpeng Lang and Hang Zhao},
  booktitle={Proceedings of The 8th Conference on Robot Learning},
  year={2025}
}

@inproceedings{CLIP,
  author       = {Alec Radford and
                  Jong Wook Kim and
                  Chris Hallacy and
                  Aditya Ramesh and
                  Gabriel Goh and
                  Sandhini Agarwal and
                  Girish Sastry and
                  Amanda Askell and
                  Pamela Mishkin and
                  Jack Clark 
                  et al.},
  title        = {Learning Transferable Visual Models From Natural Language Supervision},
  booktitle    = {Proceedings of the 38th International Conference on Machine Learning},
  year         = {2021},
}

@inproceedings{LMDrive,
  title={LMDrive: Closed-Loop End-to-End Driving with Large Language Models},
  author={Hao Shao and Yuxuan Hu and Letian Wang and Guanglu Song and Steven L. Waslander and Yu Liu and Hongsheng Li},
  booktitle={Proceedings of the IEEE/CVF Conference on Computer Vision and Pattern Recognition},
  year={2024}
}

@inproceedings{OmniDrive,
    author    = {Wang, Shihao and Yu, Zhiding and Jiang, Xiaohui and Lan, Shiyi and Shi, Min and Chang, Nadine and Kautz, Jan and Li, Ying and Alvarez, Jose M.},
    title     = {OmniDrive: A Holistic Vision-Language Dataset for Autonomous Driving with Counterfactual Reasoning},
    booktitle = {Proceedings of the IEEE/CVF Conference on Computer Vision and Pattern Recognition},
    year      = {2025},
}

@inproceedings{MA-LMM,
  author    = {He, Bo and Li, Hengduo and Jang, Young Kyun and Jia, Menglin and Cao, Xuefei and Shah, Ashish and Shrivastava, Abhinav and Lim, Ser-Nam},
  title     = {MA-LMM: Memory-Augmented Large Multimodal Model for Long-Term Video Understanding},
  booktitle = {Proceedings of the IEEE/CVF Conference on Computer Vision and Pattern Recognition},
  year      = {2024},
}

\section*{Appendix}
\begin{figure}[H]
    \centering
    \includegraphics[width=0.45\textwidth]{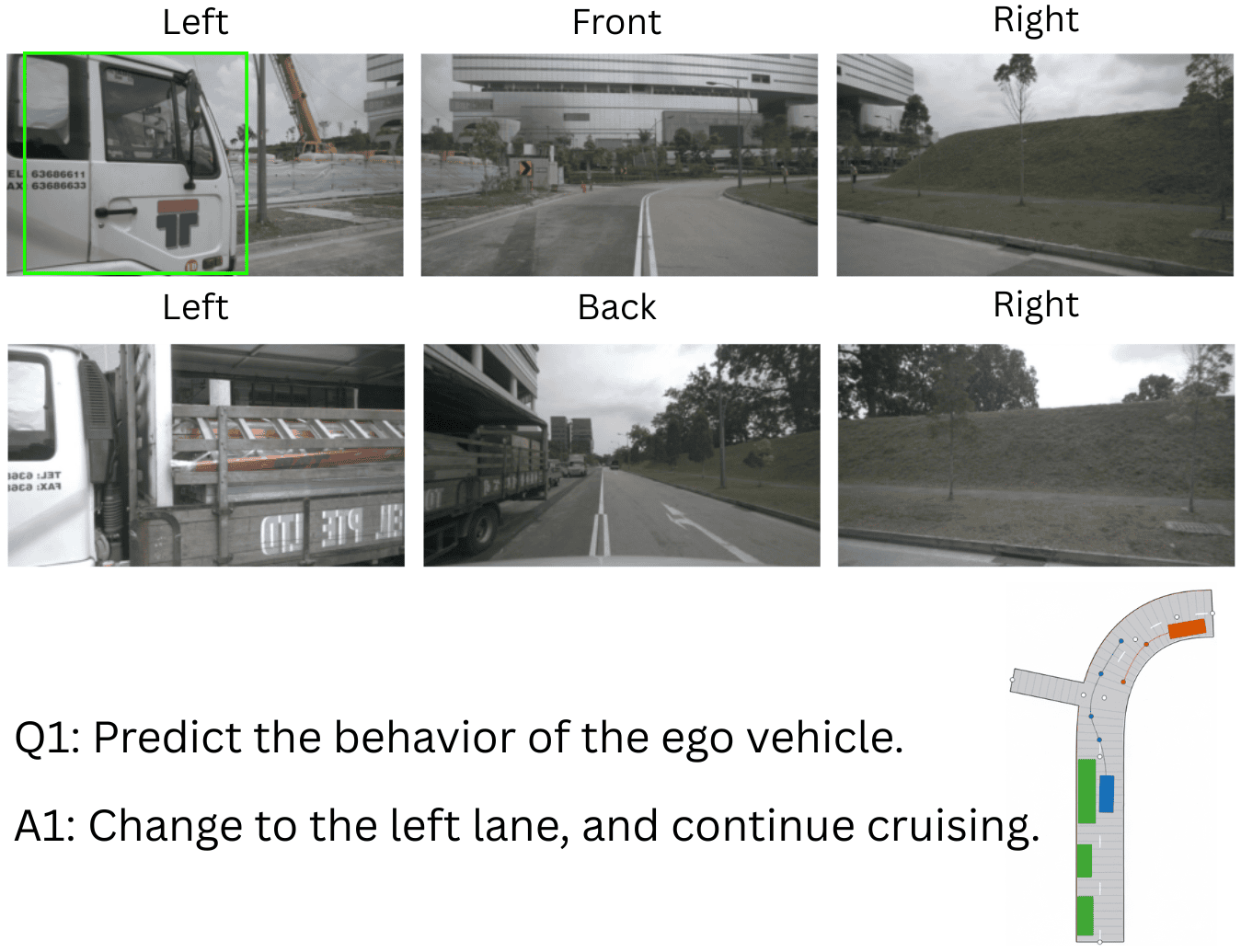}
	\caption{CommandLM behavior prediction from LiDAR and multi-view cameras; the top-down sketch shows the ego vehicle (blue), parked construction vehicles (green), and an oncoming vehicle (orange), illustrating scene context for overtake maneuver.}
    \label{fig:overtake-example-short}
\end{figure}
\begin{figure}[H]
    \centering
    \includegraphics[width=0.42\textwidth]{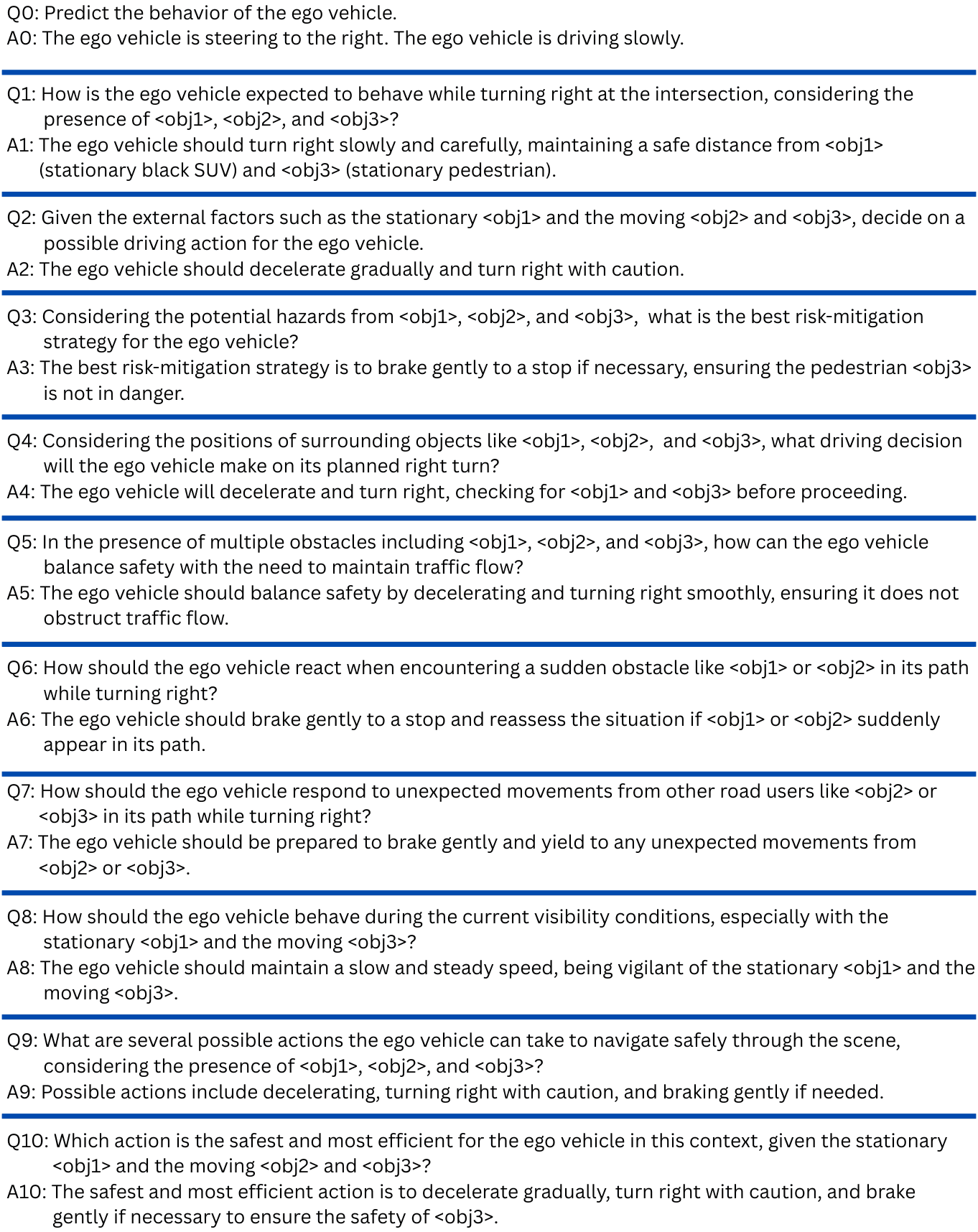}
    \caption{Example of QA pairs for a single sample, where QA0 is the original DriveLM-nuScenes pair and QA1–QA10 are part of the 40,000 newly generated CommandLM-nuScenes pairs.}
    \label{fig_additional-qas}
\end{figure}

\vspace{12pt}
\color{red}
\end{document}